\documentclass{trbunofficial}
\usepackage{multirow}
\usepackage{booktabs}
\usepackage{pdflscape}
\usepackage{float}

\makeatletter
\renewcommand{\maketitle}{%
  \TRBsetauthors
  \thispagestyle{empty}
  \begin{flushleft}
    {\large\bfseries\@title}\\
    \hfill\break%
    \@author
    \ifTRB@havecorr \textsuperscript{*}Corresponding author\\ \fi
  \end{flushleft}
  \TRB@printTitleFootnotes
  \newpage
}
\makeatother

\begin{document}


\title{3-D Emissions Mapping and Social Cost Estimation for US Domestic Aviation at West Coast Hubs}

\TRBauthor{Hesam Shafiei Nia}{Department of Civil and Environmental Engineering , University of Washington}{hesamsh@uw.edu}[Seattle, WA, 98195]  

\TRBauthor{Don MacKenzie}{Department of Civil and Environmental Engineering , University of Washington}{dwhm@uw.edu}[Seattle, WA, 98195][0000-0002-0344-2344]

\AuthorHeaders{Shafiei Nia, and MacKenzie}


\maketitle

\section{Abstract}

\hfill\break%
\noindent\textbf{Objectives:}~Existing aviation emissions inventories lack sufficient accurate trajectory data for high-resolution social cost and health impact assessment. This paper develops a three-dimensional emissions map by reconstructing flight trajectories for US west coast hubs to estimate regional environmental effects and near-airport health impacts.

\hfill\break%
\noindent\textbf{Methods:}~A physics-informed autoencoder (AE) is applied to Automatic Dependent Surveillance-Broadcast (ADS-B) trajectory records for January 2025 covering US west coast hubs. The encoder combines a Convolutional Neural Network (CNN), a Bidirectional Gated Recurrent Unit (Bi-GRU), and a three-dimensional CNN with skip connection; the decoder is a Temporal Convolutional Network (TCN). This model is benchmarked against the baseline-AE and cubic spline interpolation. Emissions are mapped via EUROCONTROL Base of Aircraft Data (BADA) performance tables and International Civil Aviation Organization Engine Emissions Databank (ICAO EEDB) emission indices, with altitude corrections via Boeing Fuel Flow Method 2 (BFFM2). Social costs are quantified for all flight phases, with health impacts assessed for Landing and Takeoff cycles within 50 km radius of each hub.

\hfill\break%
\noindent\textbf{Findings:}~The proposed AE model outperforms both a TCN-AE and cubic spline interpolation across 5\% to 50\% missing rates. Applying monetized damage costs to the emissions inventory shows that $\mathrm{NO_x}$ produces a small net cooling effect in its direct climate forcing, while accounting for 99.7\% of monetized air-quality and health cost despite being less than 0.4\% of $\mathrm{CO_2}$ by mass, making it the dominant health-cost driver.

\hfill\break%
\noindent\textbf{Novelty:}~To the authors' knowledge, this is among the first studies to combine AE-based trajectory reconstruction with separate spatial-temporal feature encoding and altitude-based emissions modeling to produce a regional aviation emissions inventory for air quality, climate impact and population exposure.

\hfill\break%
\noindent\textbf{Practical Applications:}~The resulting altitude-based emissions map and social cost estimates provide quantitative context for environmental impact assessment and near-airport health policy evaluation for US domestic aviation. 

\newpage

\section{Introduction}\label{sec:intro}
Aviation emissions can degrade local and regional air quality, and contribute to climate change in ways that vary with aircraft altitude, leading to health effects in populations near airports and regions under flight paths. Commercial aviation emissions have been estimated to be the cause of approximately 16,000 premature deaths each year globally and Landing and Takeoff (LTO) cycles are responsible for roughly a quarter of them \cite{yim2015global, grobler2019marginal}. Additionally, 90\% of the global impact per unit of fuel burn is attributed to cruise emissions, and 64\% of all damages are because of air quality impacts, while nitrogen oxides ($\mathrm{NO}_x$), carbon dioxide ($\mathrm{CO}_2$), and condensation trails (contrails), line-shaped clouds composed of ice crystals, are collectively responsible for 97\% of the total impact \cite{grobler2019marginal}. Accordingly, estimating emissions in an area to use for social costs assessments at a high spatial-temporal resolution is essential for aviation decision-making and policy design. To have accurate high-resolution emissions estimations, the approach requires complete flight trajectory data free of gaps and missing observations.

Automatic Dependent Surveillance-Broadcast (ADS-B) data is a dominant, easy to access source of high-resolution flight trajectory data. However, ADS-B data does have limitations, including systematic coverage gaps due to signal loss, receiver dropout, and incomplete network coverage that cause inconsistencies and holes in the dataset. These gaps lead to inaccuracies in emissions inventories, social costs estimates and health effects assessments~\cite{verbraak2017large}. Additionally, ADS-B is vulnerable due to significant security issues based on its open-design and lack of built-in security features \cite{ahmed2025automatic}. 

To address these ADS-B data-quality issues, whether from technical limitations or attacks, and ensure trajectory continuity, various papers have implemented data imputation methods, including different interpolation methods. First, linear and polynomial interpolation \cite{lindner2021aircraft} assumes that variations between missing points follow a linear or low-order polynomial relationship. Second, spline-based interpolations fit piecewise polynomials with continuity between missing points to ensure smoothness and accuracy of connecting the points. These have been applied to flight trajectories in three sub-methods including first-order spline interpolation \cite{sun2019wrap}, cubic-spline interpolation, which is used in our previous works \cite{shafienya20224d, shafienya20224C}, and piecewise cubic Hermite interpolation \cite{yoon2023improving}.

Traditional interpolation methods for gap-filling in ADS-B data have been widely used, but they systematically misrepresent climb and descent profiles due to fitting a smooth mathematical curve between observed endpoints with reducing the ability to capture the nonlinear relationship between altitude, speed, and engine thrust that governs real aircraft dynamics \cite{ren2026review}.

With the rapid increase of machine learning (ML) methods for various applications, and due to their advantages for data imputation by learning trajectory patterns, many studies have begun to reconstruct ADS-B trajectories more faithfully using ML models. These approaches could be ML models such as Bayesian Ridge (BR), Random Forest (RF), AdaBoost (AB), Extra Tree (ET), and the k-NN models \cite{chandar2024imputing}, recurrent neural network (RNN) sub-models such as Bidirectional Gated Recurrent Unit (Bi-GRU) \cite{galdelli2025data}, or architectures based on autoencoders \cite{ren2026review}. 

The prerequisite for an accurate estimate of social and health costs is a high-resolution emissions inventory, which can be provided for different purposes such as scientific research and climate forcing analysis \cite{wasiuk2015aircraft, wasiuk2016commercial, wilkerson2010analysis}. Emissions in aviation contribute to both air pollution and climate change, thus it is important to find out how these emissions are distributed across regions and altitudes to assess the social and health impacts caused by aviation \cite{stettler2011air}. \cite{klenner2022high} proposed an assessment model to estimate species emitted from commercial aviation in Norway. In this paper, we propose an approach that estimates emissions based on altitude to provide an emissions inventory for one month of ADS-B flight trajectories to and from the United States' west coast hubs. 

Existing literature has addressed aspects of air quality impacts, climate change, and social and health costs of the aviation sector. Air quality impacts have been estimated in \cite{grobler2019marginal, brunelle2014assessing, masiol2014aircraft} for low and high altitude flight phases. Climate impact studies estimate climate change through {$\mathrm{CO}_2$} and non-{$\mathrm{CO}_2$} effects \cite{dray2022cost,grewe2021evaluating}, including life cycle aviation atmospheric {$\mathrm{CO}_2$} emissions, contrail-cirrus and ozone formation. These types of climate impact studies focus on one year of aviation emissions for radiative forcing \cite{brasseur2016impact}, or on studying contrails \cite{singh2024understanding, teoh2024global} and aviation {$\mathrm{NO}_x$} emissions \cite{kohler2008impact,harlass2024measurement}. Authors in \cite{hsu2013contributions,zhu2011aircraft} illustrate a correlation between aircraft operations and pollutant concentration near airports. Health impacts of aviation sector have been studied for near airport, regional, local scales in different countries \cite{yim2015global,zhang2023increased} to inform policy-makers and regulatory efforts. 

The above referenced studies have advanced understanding of aviation emissions, air quality, climate, and health impacts specifically using ADS-B trajectory data. However, most rely on interpolation-based approaches for handling missing data or irregular trajectory observations. These methods fit smooth mathematical curves between observed endpoints without accounting for the nonlinear relationship between altitude, speed, and engine thrust that drive the aircraft dynamics \cite{klenner2022high, teoh2024high}. By learning trajectory patterns directly from complete historical flights, ML-based imputation approaches offer a more physically constrained reconstruction of missing segments. Additionally, studies addressing climate-scale impacts and those focused on near-airport health effects have largely been conducted independently, with few works integrating both into their framework. This paper addresses both limitations by combining an autoencoder-based trajectory data reconstruction model with emissions inventory that supports simultaneous estimation of climate and near-airport health costs.

The outcome of this paper is an approach whose results decision-makers and policy-makers can use to reduce climate, air quality and health impacts of aviation by operational enhancements and policy introduction. As such our results aim at reducing aviation impacts on environment and human health.

\section{Methodology}
The proposed approach estimates emissions inventory, social costs and health effects based on altitude. First, a data imputation model is applied to ADS-B flight trajectories to correct missing data points and segments. Second, the inventory of emissions estimations is achieved using the Eurocontrol Base of Aircraft Data (BADA) aircraft-type specific model \cite{nuic2010bada, eurocontrol}, and the International Civil Aviation Organization Engine Emissions Databank (ICAO EEDB) data set \cite{icao_engine_emissions_databank} along with Boeing Fuel Flow Method 2 (BFFM2) \cite{dubois2006fuel} was used to correct altitude-dependent emission indices. Third, using the produced emissions inventory, and estimates of the external costs of emissions, the proposed framework estimates social cost and health impact between and around the hubs. 

This methodology provides a basis for evaluating emissions-related policy options for US domestic flights, with a focus on the west coast region. The remainder of this section describes each step in more detail. 

\subsection{Data and processing}
ADS-B trajectory records were extracted from OpenSky Network \cite{schafer2014bringing}, from which 20,134 flight trajectories for January 2025 were identified for the three west coast hubs, including, Seattle-Tacoma International Airport (KSEA), Los Angeles International Airport (KLAX), and San Francisco International Airport (KSFO) in the United States. The ADS-B dataset includes time, longitude, latitude, altitude, groundspeed, vertical rate, track, onground, callsign, icao24, phase, \text{trajectory\_id}, and departure airport.

Prior to model input, all trajectory records are resampled to a uniform 30 second temporal resolution and padded or truncated to a fixed length of 320 timesteps, corresponding to 160 minutes. A fixed input length is required because the encoder compresses each flight into a single fixed size latent vector and processes batches on the GPU, which require every input to share the same shape; padding and truncation reconcile this fixed-length requirement with the variable duration of real flights. The window accommodates the longest flight considering KLAX-KSEA at approximately 136 minutes with additional margin for weather, Air Traffic Control (ATC) management, and contingencies. This resampling applies only to the trajectory reconstruction step; once imputation is complete, reconstructed values are interpolated back to the original 1 Hz resolution before being passed to the emissions estimation pipeline. The downsampling therefore has no effect on the emissions inventory or social cost calculations.

The next step of the preprocessing  is to split the flight trajectories into two groups of complete and incomplete trajectories to take care of missing data. The filtering rule is that a trajectory is considered complete only if its largest gap is smaller than or equal to 36 seconds referring to Federal Aviation Administration (FAA) \cite{faa_papr_users_guide_2020}, otherwise it is incomplete. The analysis shows that 53.4\% of flight trajectories need imputation for 36 s threshold. The next section is dedicated to the proposed Autoencoder model to reconstruct flight trajectories before applying the emissions estimation model.

\subsection{Trajectory imputation}
A trajectory is a multivariate time series of surveillance data broadcast by transponders to ground stations every second. This causes flaws in data points and may result in gaps as blocks of consecutive missing timesteps. To handle missing data, especially block-missing data, which could cause the loss of all local context, we propose an autoencoder-based data imputation model to reconstruct incomplete trajectories in a physically plausible way, by learning patterns from complete trajectories and checking against fundamental physical constraints, so that downstream emissions and social cost estimations are accurate and realistic.

\subsubsection{Proposed architecture (CBiG3D-AE)}
The proposed architecture is a two-branch autoencoder with a skip connection and a Temporal Convolutional Network (TCN) decoder, the parts of which are described below.

\subsubsection{Encoder}
The proposed autoencoder is designed to look at trajectories in as many ways as possible. Accordingly, before inputting the data features to the model, we separate spatial (i.e., latitude, longitude, and altitude) and temporal (i.e., timesteps, ground speed, track/heading, vertical rate) features to encode each flight trajectory through two parallel branches, one hybrid and one Convolutional Neural Network (CNN)-based model. Each branch is designed to capture a different aspect of the motion, and combine the output into a single representation for the purpose of increasing the accuracy in detecting flight patterns. 

The first branch contains two bodies. The first body is a one-dimensional CNN applied to extract spatial features, consisting of three convolutional blocks of increasing width, followed by normalization, an activation function, and a pooling step to help with compressing the sequence. Due to the ability of convolutional networks in responding to local patterns, wherever they occur, they are the suitable option to capture the overall geometry of a flight. The second body, which works in parallel with CNN as a hybrid model, is a bidirectional recurrent network, specifically a Bi-GRU, applied to the temporal features, because it reads the sequence in both the forward and backward directions, which enables each hidden state to integrate features from both earlier and later time sequences in the input window. While the convolutional body sees local shape, the recurrent body captures dynamic and momentum, which means the tendency of an aircraft to keep its move when climbing or descending at a consistent rate. Using a bidirectional recurrent body is valuable to the model to see the evidence in either side of a gap when inferring what happened inside it. The 30 seconds window helps keep computation expenses reasonable while prevents loss of information by BiGRU by reducing the numbers of sequences. 

The second branch captures how position and time interact, which the first branch bodies treat only separately. The 3-Dimensional Convolutional Neural Network (3D-CNN) considers a flight trajectory as a stack of two-dimensional maps, one per timestamp, in which the aircraft's location is represented as a point of intensity. Then, the model processes this stack, and learns patterns of time and space at once. This branch encodes the aircraft's spatial-temporal evolution. This information complements the outcome from the first branch including route shape from the first body and the temporal dynamics from the second. 

The outputs of the two branches and three bodies are concatenated and passed through a small fusion network that compresses them into a latent space represented by a single fixed-size vector. This vector is a representative pattern of the entire flight. By using both branches, and all three bodies at once, the encoder captures a pattern that could not have been detected this accurately if we had used each body alone.

We added a skip connection to the encoder model to enhance the model's generalization performance \cite{oyedotun2022everyone}. The skip connection is implemented as a path that carries information from the input, without downsampling, directly to the later stage of decoding by skipping the layers in between. The reason for using a skip connection is that the encoder branches compress the whole flight into a 256-dimensional latent vector, which causes the loss of the fine, local detail the model needs to fill the gap.  

\begin{figure}[htbp]
  \centering
  \includegraphics[width=1\linewidth]{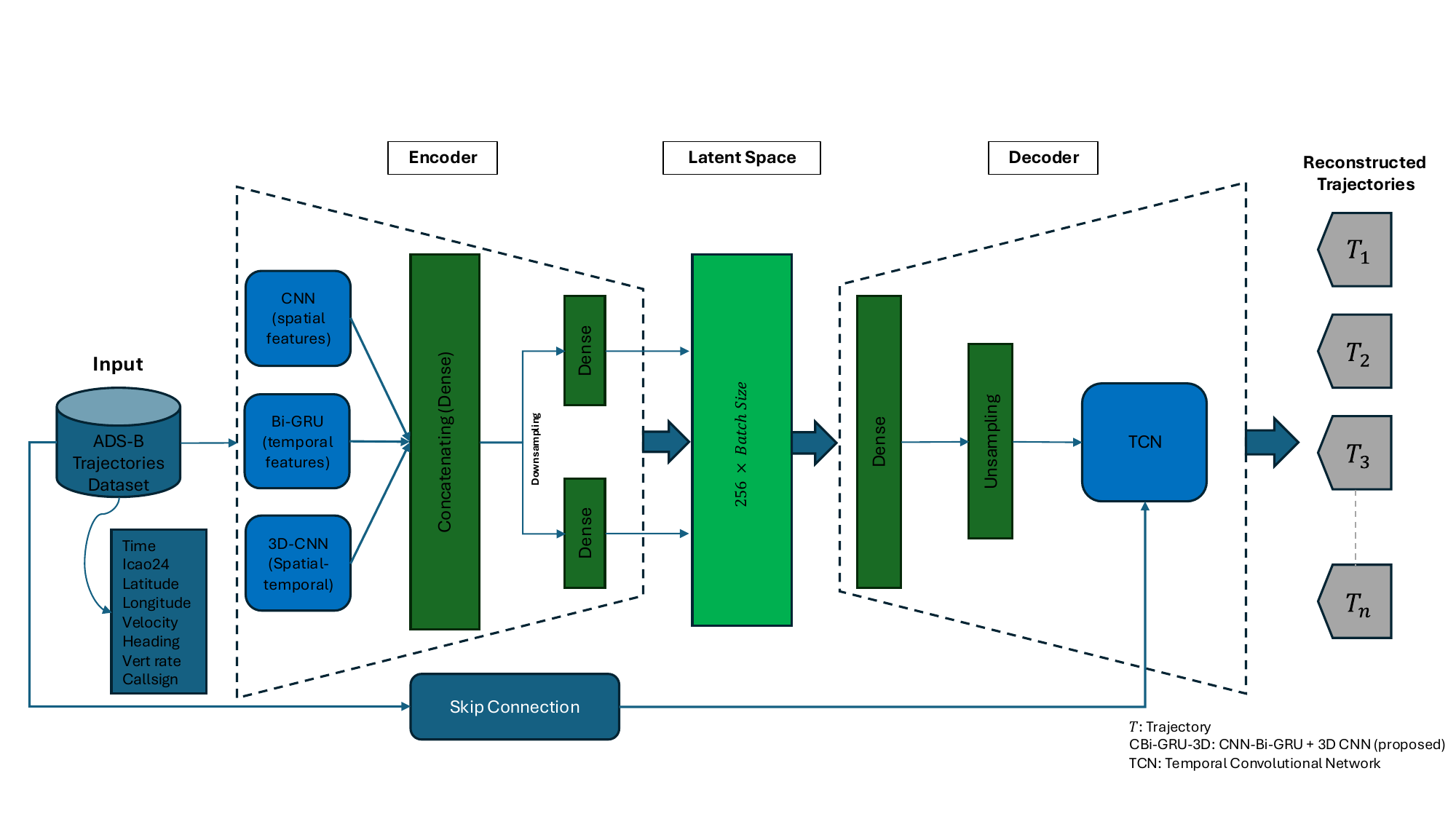}
  \caption{Proposed deep autoencoder architecture of the CBiG3D model.}\label{fig:proposedAE}
\end{figure}

This added skip branch architecture applies convolutions with a preserved time dimension, so that every vector's features remain assigned to their timestamp. The dilated kernels project information from the observed neighbors on each side of the gap into the gap position. In other words, skip connections help preserve spatial accuracy by bringing forward detailed features from earlier layers. The decoder begins to reconstruct the trajectory through unsampling, at each level it combines decoder features with corresponding encoder features using skip connections to ensure that reconstructions stay related to the actual data around each gap \cite{geeksforgeeks_unet_2025}. The proposed deep autoencoder model is shown in Figure~\ref{fig:proposedAE}.

\subsubsection{Decoder}
The encoded latent vector is decoded into a full trajectory using a TCN. During decoding, a linear layer first projects the latent vector into a sequence-shaped feature map, which then passes through multiple convolutional blocks whose receptive fields grow rapidly with depth using dilation. This causes the receptive field to expand exponentially, allowing the decoder to reconstruct long gaps thoroughly rather than only short ones. Because the convolutions look both backward and forward in the sequence, every reconstructed point is inferred by considering both sides of the gap. The reason to choose TCN over Long Short-Term Memory (LSTM) or GRU for the decoding is its ability to process the whole sequence in parallel, which increases the speed and accuracy of the results and ensures stable behavior over long spans.

\subsubsection{Training objectives}
The model is trained using complete flight trajectories with no gap or missing data points without manual labels; for this purpose we used 27,500 complete flight trajectories using only coordinates and kinematic features to let the model learn what a flyable flight trajectory looks like. These training trajectories are drawn from a broader historical flight-trajectory corpus, held separate from the January 2025 west-coast dataset used for the emissions case study; the latter is reserved entirely for evaluation and is not used during training. The AE model deliberately corrupts part of a complete trajectory and asks the network to reconstruct the original, using the uncorrupted flight as ground truth. The corruption modes, gap lengths, starting positions, and number of realizations per trajectory are described below.

Three corruption modes are applied, selected independently at random for each training sample with probabilities 40\%, 40\%, and 20\%, respectively: (1) \textit{scattered masking}, in which a fixed 20\% of timesteps are selected uniformly at random (without replacement) and set to missing, simulating sparse independent dropouts; (2) \textit{contiguous block masking}, in which a single gap is placed at a uniformly random starting position, with its length drawn uniformly at random between 5\% and an epoch-dependent maximum, simulating the block gaps characteristic of real ADS-B dropouts; and (3) \textit{spike injection}, in which 5\% of timesteps are perturbed with implausible altitude values while the target remains unchanged, training the model to suppress sensor artifacts rather than treat them as signal. For contiguous gaps, the maximum allowed gap length follows a curriculum that widens from 15\% of the sequence length in the first training epoch to 50\% by the final epoch, exposing the model to progressively harder reconstruction tasks as training proceeds. Because corruption is resampled independently every epoch, each training trajectory is exposed to a new, randomly realized corruption pattern at every one of the 150 training epochs, rather than a fixed, pre-generated set of augmented copies. These models were used only during training to allow the model to learn missing data patterns. Validation uses a single fixed 20\% contiguous gap as the corruption type most representative of real ADS-B dropouts, for every trajectory, rather than the training curriculum, so that checkpoint selection is based on a consistent task across epochs.

Reconstructed trajectories must also be physically flyable, not just close to the target in pointwise error, due to the downstream emissions inventory's direct reliance on plausible altitude and speed profiles. The loss function combining these objectives is defined below. 


The total training loss combines a reconstruction term and physics-informed regularizers to minimize a composite loss over masked timesteps:

\begin{equation}
  \mathcal{L}= \mathcal{L}_{\mathrm{rec}}
              + \lambda_{\mathrm{lat}}\,\mathcal{L}_{\mathrm{lat}}
              + \lambda_{v}\,\mathcal{L}_{v}
              + \lambda_{s}\,\mathcal{L}_{s}
              + \lambda_{b}\,\mathcal{L}_{b}
              + \lambda_{gs}\,\mathcal{L}_{gs}
  \label{eq:base_loss}
\end{equation}

\paragraph{Reconstruction loss $\mathcal{L}_{\mathrm{rec}}$.}
Penalizes error only inside the masked region $\mathcal{M}$, which is the
synthetically removed segment:
\begin{equation}
  \mathcal{L}_{\mathrm{rec}}
    = \frac{1}{|\mathcal{M}|}
      \sum_{t \in \mathcal{M}} \bigl(F_r^{(t)} - F_t^{(t)}\bigr)^2
  \label{eq:loss_rec}
\end{equation}
where $F_r^{(t)}$ is the reconstructed value and $F_t^{(t)}$ is the
actual value at masked step $t$.

\paragraph{Latent regularizer $\mathcal{L}_{\mathrm{lat}}$.}
Constrains the $d$-dimensional latent vector $\mathbf{z}$ to prevent
memorization of individual flights and push the model to learn of general
trajectory patterns:
\begin{equation}
  \mathcal{L}_{\mathrm{lat}} = \|\mathbf{z}\|^2
  \label{eq:loss_lat}
\end{equation}

\paragraph{Vertical-rate consistency $\mathcal{L}_{v}$.}
Enforces kinematic consistency between reconstructed altitude and
vertical rate over each time step $\Delta t$:
\begin{equation}
  \mathcal{L}_{v}
    = \frac{1}{|\mathcal{M}|}
      \sum_{t \in \mathcal{M}}
      \left(\frac{\Delta h_r^{(t)}}{\Delta t} - v_r^{(t)}\right)^2
  \label{eq:loss_vrate}
\end{equation}
where $\Delta h_r^{(t)}$ is the reconstructed altitude change and
$v_r^{(t)}$ is the reconstructed vertical rate.

\paragraph{Smoothness penalty $\mathcal{L}_{s}$.}
Penalizes large step to step changes in vertical rate, enforcing
physically plausible climb and descent profiles:
\begin{equation}
  \mathcal{L}_{s}
    = \frac{1}{|\mathcal{M}|-1}
      \sum_{t \in \mathcal{M}}
      \bigl(v_r^{(t)} - v_r^{(t-1)}\bigr)^2
  \label{eq:loss_smooth}
\end{equation}

\paragraph{Altitude bounds $\mathcal{L}_{b}$.}
Applies a one-sided penalty when the reconstructed altitude violates
the operational envelope, with zero penalty inside the plausible range
$[h_{\min},\, h_{\max}]$:
\begin{equation}
  \mathcal{L}_{b}
    = \frac{1}{|\mathcal{M}|}
      \sum_{t \in \mathcal{M}}
      \Bigl[
        \max\!\bigl(0,\; h_r^{(t)} - h_{\max}\bigr)^2
      + \max\!\bigl(0,\; h_{\min} - h_r^{(t)}\bigr)^2
      \Bigr]
  \label{eq:loss_bounds}
\end{equation}
where $h_{\max} \approx 13{,}000$~m is the commercial service ceiling, following \cite{aeroclass2021howhigh}, and $h_{\min}$ is terrain elevation at the waypoint.
Weights $\lambda_{\mathrm{lat}} = 0.0001 $, $\lambda_{v} = 0.1$,
$\lambda_{s} = 0.15$, and $\lambda_{b} = 0.05$ were set as fixed parameters based on the relative scale of each physics attribute.

\paragraph{Groundspeed smoothness $\mathcal{L}_{gs}$.}
Penalizes large step to step changes in reconstructed groundspeed over the
full sequence, discouraging implausible instantaneous speed jumps:
\begin{equation}
  \mathcal{L}_{gs}
    = \frac{1}{T-1}
      \sum_{t=1}^{T-1}
      \left(\frac{g_r^{(t+1)} - g_r^{(t)}}{g_{\max}}\right)^2
  \label{eq:loss_gs}
\end{equation}
where $g_r^{(t)}$ is the reconstructed groundspeed at step $t$ and
$g_{\max} \approx 350$~m/s is the groundspeed normalization ceiling, to keep the penalty on a comparable level to the other implied physics terms.

The system we used for training the models is a University of Washington Hyak computing cluster that uses a NVIDIA L40 GPU with 140 GB of RAM, running for 13 hours. The models are optimized with the decoupled weight decay (AdamW) optimizer and standard regularizer, and we used fixed 150 epochs for each model. 

\subsubsection{Data imputation}
After the model is trained, it is applied to the incomplete flights. Each flight is passed through the model to fill in its missing points and blocks, and the model reconstructs the full sequence. To get the best results out of the model for reconstructed trajectories, each flight passes to the model multiple times. It is important to note that the actual observed points are always kept exactly as recorded, and only the missing points/blocks get their filled points. At the last stage of the trajectory imputation model, the imputed flight trajectories are merged with the complete trajectories to make a full-resolution trajectory dataset that is passed to the next stage of the framework to estimate the emissions. The results of 36 second gap threshold show that 10,751 of 20,134 west coast flight trajectories in January 2025, about 53.4\% with the average within flight missing fraction of 13.5\%, require data imputation.  

\subsubsection{Baseline AE model}
To evaluate the proposed autoencoder model, we compare it against two types of approaches, a deep learning model that does not have the separate spatial-temporal contribution and added skip connection of the proposed multi-branch AE model, and a conventional cubic spline interpolation method that represents current practice. 

\subsubsection{TCN-AE}
To compare the proposed autoencoder with a model that does not have the separate spatial-temporal features contribution of our model we used a TCN-AE model with TCN as both encoder and decoder. It uses the same training procedure, and the same masking approach self-supervised as our proposed model, but replaces our proposed two branches encoder with a TCN encoder taking all spatial-temporal features together. This provides a control baseline that shows any difference in performance between the proposed model and the TCN-AE which is attributable specifically to the spatial-temporal two branch design of the proposed AE encoder. The parameter counts between the proposed models and baseline model are comparable, so the comparison is not confounded by model capacity.

\subsubsection{Interpolation methods}
This section of comparison models contains one conventional interpolation method widely used by researchers for their trajectory related studies and missing data imputation. The method is as follows. 

Cubic spline interpolation fits a smooth piecewise-cubic curve through the observed points. It produces smoother trajectories compared to the linear interpolation but can generate physically implausible paths. This method cannot detect noise and may interpolate to an out of range altitude, which causes errors that can affect the accuracy of our emissions estimations.

The trained autoencoder reconstructs missing data in a way that is consistent with real-world flight dynamics/behavior. Comparing learning-based models against traditional imputation methods illustrates their practical value in missing data imputation.

\subsection{Evaluation}
All imputation methods are evaluated on the same held-out trajectories, each given a single contiguous synthetic gap of known ground truth, with length drawn uniformly at random between 5\% and 20\% of the trajectory length, matching the block-gap pattern characteristic of real ADS-B dropouts. This is also the harder to handle between corruption types used during training for interpolation-based baselines to reconstruct, making it the more informative test of the proposed model's advantage; scattered single-point dropouts, the easier condition, are therefore not reported separately. Spike injection targets implausible sensor artifacts rather than missing data, thus it is not reported here. Evaluation metrics are Mean Absolute Error (MAE) and Root Mean Squared Error (RMSE) for each trajectory feature, together with a position error in dynamic time distance (km) that measures overall trajectory-shape accuracy. Each metric is computed over the $n$ masked timesteps $\mathcal{M}$~, Eq.~\ref{eq:mae}~and~\ref{eq:rmse}:

\begin{equation}
  MAE = \frac{1}{n}\sum_{t=1}^{n}\left|F_r - F_t\right|
  \label{eq:mae}
\end{equation}

\begin{equation}
  RMSE = \sqrt{\frac{1}{n}\sum_{t=1}^{n}\left(F_r - F_t\right)^2}
  \label{eq:rmse}
\end{equation}
where $F_r$ is the reconstructed value from the autoencoder and $F_t$ is
the true observed value at the masked timestep.


\subsection{Emissions estimation framework}
The proposed framework estimates emissions species across altitudes geospatially and temporally utilizing Eurocontrol BADA 3.16 performance tables, using trajectories instantaneous pressure altitude, calibrated True Air Speed (TAS), and vertical rate for operating aircraft types fuel flow calculation with ICAO EEDB emission indices as the sea-level, static reference for the nearest matching thrust setting, corrected to actual flight conditions at every timestep via BFFM2. This correction is applied uniformly across Landing and Takeoff (LTO) and non-LTO flight phases alike, since aircraft speed during the LTO cycle differs meaningfully from the near-static reference conditions under which the ICAO values were certified, making the correction non-trivial even at low altitude. BFFM2 was initially developed by Boeing to adapt emission factors and fuel flow to atmospheric conditions \cite{schaefer2013overview}.

\begin{figure}[htbp]
  \centering
  \includegraphics[width=0.7\linewidth]{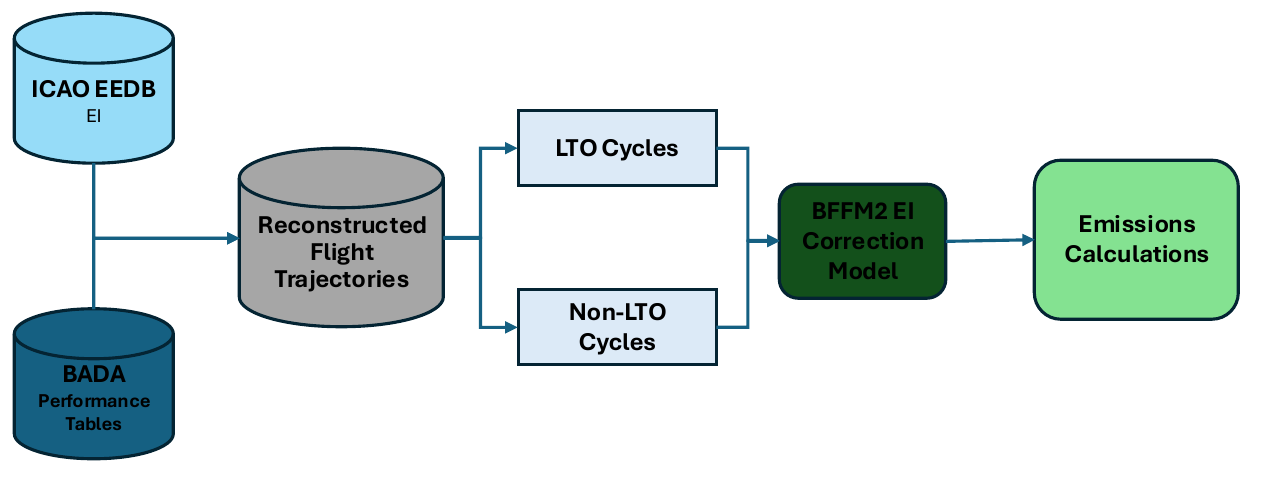}
  \caption{Emissions estimation approach schematics.}\label{fig:emissions_estimations}
\end{figure}

To use BADA performance attributes, we assumed constant aircraft weights relying on BADA-defined values. The emissions estimation process is shown in Figure~\ref{fig:emissions_estimations}.

With fuel flow calculated from the previous step, emissions could be assessed. Emissions per trajectory across altitudes at time ($t$) are the product of fuel consumption ($kg/s$) and EI based on altitude given in kg emissions per kg of fuel consumed. The total mass of each emissions species $i$ during a flight across altitudes is calculated using Eq.~\ref{eq:emission}. 

\begin{align}
  &  e_i = \sum_{t} EI_{i,t} \cdot FF_t \cdot \Delta t \label{eq:emission}
\end{align}

Where $e_i$ is the total mass of emissions species $i$ (g), $EI_{i,t}$ is the emission index for species $i$ at timestep $t$ (g/kg fuel), $FF_t$ is the fuel flow rate per aircraft at timestep $t$ (kg/s), and $\Delta t$ is the time gap between two timesteps (s).

As demonstrated in Figure~\ref{fig:emissions_estimations}, emissions species including $\mathrm{CO}_2$ for climate effect, $\mathrm{NO}_x$, $\mathrm{HC}$ and $\mathrm{CO}$ for air quality analysis, and Non-volatile particulate matter ($nvPM$) for health effects have been estimated using the proposed approach with a coverage of the west coast area, and near hubs locations for LTO and non-LTO cycles, and LTO cycles respectively.

\subsection{Social cost and health impact framework}
Social costs are external costs which are directly depend on emissions species and their damages emitted by flights across their trajectories. In this paper, we assess costs of these damages on climate, air quality and human health using estimated emissions map for the study area. 

Aviation's contribution to climate change has been studied by \cite{marais2008assessing, mahashabde2011assessing, wolfe2015aviation} by computing probabilistic estimates of aviation's climate using a quasi-Monte Carlo method considering multiple economies and policy scenarios. In this study, we used the constant number of 190 USD per metric tone of $\mathrm{CO}_2$ from the U.S. Environmental Protection Agency (EPA) report on social costs of greenhouse gases \cite{epa_scghg_2023} to estimate a tangible cost of this pollutant's damage on climate change by multiplying the number by calculated emissions from the emissions inventory.

Non-$\mathrm{CO}_2$ pollutants ($\mathrm{NO}_x$, HC, CO) are priced using phase-specific marginal climate and air-quality damage costs for aviation emissions from \citet{grobler2019marginal}, also for nvPM climate cost is \$18{,}000/t (LTO) and \$52{,}000/t (cruise), which we assumed negligible;they compute climate impacts using a reduced-order climate model and air-quality/health impacts using the adjoint of the GEOS-Chem chemistry-transport model combined with concentration-response functions and the value of a statistical life (VSL). This is a direct per pollutant marginal damage cost approach rather than a Global Warming Potential (GWP)-based conversion to $\mathrm{CO}_2$-equivalent mass priced at the $\mathrm{CO}_2$ Social Cost of Carbon (SCC). \citet{grobler2019marginal} report $\mathrm{NO}_x$'s direct climate effect as a small net cooling, since $\mathrm{NO}_x$-driven methane destruction outweighs short-lived tropospheric ozone warming at a 3\% discount rate, while its air-quality/health cost is two orders of magnitude larger and dominates its total impact. Values are applied separately for the landing-and-takeoff (LTO, $<$914~m) and non-LTO (cruise) phases, consistent with the phase boundary used elsewhere in this paper.

Air quality and health impact has been assessed in amount of emitted $\mathrm{NO}_x$, $\mathrm{CO}$, $\mathrm{HC}$, and $nvPM$ and showing their damages by amount of money and premature deaths causes by responsible species. The values and their references are provided in Table~\ref{tab:damage_costs} for both climate and air quality/health damages. 

\begin{table}[htbp]
  \centering
  \footnotesize
  \setlength{\tabcolsep}{6pt}
  \renewcommand{\arraystretch}{0.85}
  \caption{Unit damage costs used to monetize climate and near-airport air quality/health impacts.}
  \label{tab:damage_costs}
  \resizebox{\linewidth}{!}{%
  \begin{tabular}{ll rr l}
    \hline
    & & \multicolumn{2}{c}{Value (USD/t)} & \\
    \cmidrule(lr){3-4}
    Category & Species & LTO & Cruise & Reference \\
    \hline
    \multirow{2}{*}{Climate (all altitudes, all flight phases)}
     & $\mathrm{CO}_2$ & 190     & 190     & \cite{epa_scghg_2023} \\
     & $\mathrm{NO}_x$ & $-$590  & $-$940  & \cite{grobler2019marginal} \\
    \hline
    \multirow{4}{*}{\shortstack[l]{Air quality \& Near-airport health\\(LTO $\leq$50 km radius)}}
     & $\mathrm{NO}_x$$^b$          & 20,000  & 24,000  & \cite{grobler2019marginal} \\
     & HC (NMVOC)$^a$           & 19,000  & --  & \cite{grobler2019marginal} \\
     & CO                       & 520     & --  & \cite{grobler2019marginal} \\
     & $\mathrm{PM}_{2.5}$      & 600,000 & --  & \cite{brunelle2014assessing} \\
    \hline
  \end{tabular}}
  \vspace{2pt}
  {\scriptsize $^a$ NMVOC: Non-Methane Volatile Organic Compounds. $^b$ $\mathrm{NO}_x$ uses USA-specific marginal costs from \citet{grobler2019marginal} for both phases.\par}
\end{table}

\section{Results}
The previous sections describe the methodology for obtaining accurate data on both emissions inventory and social costs, which depend on each other. Here we illustrate the results for the data imputation methods and emissions map to quantify the social cost and health impact associated with relevant emissions species.

\subsection{Imputation accuracy}
To verify the accuracy of our proposed AE-based model and compare it with other methods and architectures two other models were trained for the dataset that demonstrate the efficiency of the model in Table~\ref{tab:imputation_comparison}. Overall MAE and RMSE aggregate error across all eight normalized trajectory features at gap positions, computed as an unweighted mean (MAE) or pooled route-mean-squared (RMSE) over normalized error of all features combined. Per-feature validation errors are reported separately in Table~\ref{tab:imputation_comparison} for interoperability. 
Because six of the eight features are normalized to $[0,1]$ and the two heading components to $[-1,1]$, this aggregate is an approximate, equally-weighted summary rather than a physically-weighted composite reflecting each feature's downstream importance to emissions accuracy; this is the reason we report altitude and vertical-rate accuracy separately as the metrics most directly relevant to this study's fuel flow and emissions calculations.

Accordingly, the proposed AE architecture achieves the best overall accuracy in reconstructing trajectories over the baseline-AE and traditional cubic spline interpolation methods. Also, the proposed approach outperforms the other two on vertical-rate and altitude which are most directly tied to fuel flow and emissions estimations. The results show that the cubic spline method outperforms our proposed method and the baseline-AE on raw position and ground speed, because of trajectories being near linear in these quantities for short and moderate flights we studied in this paper; This shows the strength of local interpolation in these situations rather than a weakness of the proposed model. The comparison between the baseline-AE and our proposed AE model demonstrates the superiority of our model in handling flight trajectory data imputation for different distributions for training and test data.  

\begin{table*}[htbp]
\centering
\caption{Imputation accuracy (mean $\pm$ SD), by model and trajectory attribute.}
\label{tab:imputation_comparison}
\renewcommand{\arraystretch}{0.85}
\footnotesize
\begin{tabular}{llccc}
\toprule
Attribute & Metric & Proposed AE\textsuperscript{a} & TCN-AE & Cubic Spline \\
\midrule
\multirow{2}{*}{Overall}
 & MAE  & \textbf{0.061$\pm$0.056} & 0.184$\pm$0.143 & 2.603$\pm$53.833 \\
 & RMSE & \textbf{0.139$\pm$0.093} & 0.274$\pm$0.183 & 10.488$\pm$221.482 \\
\midrule
\multirow{2}{*}{Latitude (deg)}
 & MAE  & 0.416$\pm$0.929 & 2.896$\pm$3.140 & \textbf{0.238$\pm$1.118} \\
 & RMSE & 0.507$\pm$1.131 & 3.090$\pm$3.300 & \textbf{0.298$\pm$1.475} \\
\midrule
\multirow{2}{*}{Longitude (deg)}
 & MAE  & 0.274$\pm$0.467 & 3.000$\pm$2.894 & \textbf{0.181$\pm$0.829} \\
 & RMSE & 0.326$\pm$0.582 & 3.201$\pm$3.008 & \textbf{0.232$\pm$1.146} \\
\midrule
\multirow{2}{*}{Position (km)}
 & MAE  & 58.02$\pm$107.97 & 457.02$\pm$397.26 & \textbf{33.97$\pm$133.00} \\
 & RMSE & 69.17$\pm$131.39 & 475.06$\pm$410.52 & \textbf{42.04$\pm$170.73} \\
\midrule
\multirow{2}{*}{Altitude (m)}
 & MAE  & \textbf{311.4$\pm$426.2} & 2529.3$\pm$3089.4 & 929.2$\pm$14856.5 \\
 & RMSE & \textbf{357.7$\pm$502.8} & 2639.8$\pm$3123.2 & 1323.6$\pm$22152.7 \\
\midrule
\multirow{2}{*}{Groundspeed (m/s)}
 & MAE  & 12.06$\pm$39.07 & 26.20$\pm$41.43 & \textbf{3.41$\pm$15.93} \\
 & RMSE & 15.15$\pm$45.87 & 28.82$\pm$43.36 & \textbf{4.18$\pm$18.77} \\
\midrule
\multirow{2}{*}{Vert. rate (m/s)}
 & MAE  & \textbf{12.42$\pm$9.96} & 13.82$\pm$9.33 & 1197.88$\pm$25148.74 \\
 & RMSE & \textbf{16.13$\pm$11.37} & 18.01$\pm$10.45 & 1767.95$\pm$37579.71 \\
\bottomrule
\end{tabular}
\vspace{2pt}

{\scriptsize\textsuperscript{a}CNN-GRU + 3D-CNN encoder with skip connection, and TCN decoder. Bold indicates the best-performing method for each row.\par}
\end{table*}

To evaluate the presented AE model's performance, the imputation process was repeated at fixed synthetic gap rates ranging from 5\% to 50\%, plus one additional scenario emulating the real-data average missing fraction (13.5\%) by introducing a synthetic gap of that size into the same held-out complete trajectories, since genuinely incomplete flights have no known ground truth against which to measure accuracy. Results are reported in Table~\ref{tab:sensitivity}. The results illustrate that, our model leads in overall MAE and RMSE for all scenarios, and in position and groundspeed accuracy outperforms the cubic spline interpolation for higher missing rates. This shows the capability of the proposed model in handling more corrupted/spoofed trajectories compared to the other models. With reconstructed trajectories available, we next estimate altitude-based emissions using BADA and ICAO EEDB.  

\begin{table*}[htbp]
\centering
\caption{Imputation MAE and RMSE (mean $\pm$ SD) Synthetic and real missing-data rate analysis.}
\label{tab:sensitivity}
\resizebox{\textwidth}{!}{%
\footnotesize
\begin{tabular}{llcccccccc}
\toprule
& & \multicolumn{2}{c}{Overall} & \multicolumn{2}{c}{Position (km)} & \multicolumn{2}{c}{Altitude (m)} & \multicolumn{2}{c}{Groundspeed (m/s)} \\
\cmidrule(lr){3-4} \cmidrule(lr){5-6} \cmidrule(lr){7-8} \cmidrule(lr){9-10}
Rate & Model & MAE & RMSE & MAE & RMSE & MAE & RMSE & MAE & RMSE \\
\midrule
\multirow{3}{*}{5\%}
 & Proposed AE   & \textbf{0.050$\pm$0.063} & \textbf{0.116$\pm$0.103} & 44.07$\pm$141.54 & 47.76$\pm$149.04 & \textbf{135.7$\pm$266.5} & \textbf{150.8$\pm$301.0} & 14.11$\pm$51.56 & 15.39$\pm$53.75 \\
 & TCN-AE        & 0.185$\pm$0.148 & 0.274$\pm$0.193 & 464.46$\pm$443.42 & 471.80$\pm$447.84 & 2388.8$\pm$3066.9 & 2423.7$\pm$3070.4 & 26.50$\pm$43.80 & 27.71$\pm$44.48 \\
 & Cubic Spline  & 0.111$\pm$0.317 & 0.314$\pm$0.869 & \textbf{13.42$\pm$73.89} & \textbf{16.61$\pm$96.41} & 283.1$\pm$4121.7 & 385.0$\pm$5982.6 & \textbf{1.87$\pm$7.83} & \textbf{2.23$\pm$9.48} \\
\midrule
\multirow{3}{*}{10\%}
 & Proposed AE   & \textbf{0.058$\pm$0.055} & \textbf{0.134$\pm$0.094} & 51.03$\pm$110.73 & 59.77$\pm$132.62 & \textbf{252.0$\pm$373.1} & \textbf{289.5$\pm$450.7} & 12.06$\pm$40.08 & 14.77$\pm$46.17 \\
 & TCN-AE        & 0.184$\pm$0.145 & 0.275$\pm$0.186 & 455.53$\pm$403.09 & 471.35$\pm$414.90 & 2499.1$\pm$3115.9 & 2585.0$\pm$3138.1 & 26.69$\pm$43.00 & 28.83$\pm$44.46 \\
 & Cubic Spline  & 0.368$\pm$5.906 & 1.209$\pm$21.923 & \textbf{38.53$\pm$276.62} & \textbf{47.01$\pm$322.32} & 553.1$\pm$6828.1 & 758.1$\pm$10017.2 & \textbf{3.41$\pm$25.24} & \textbf{4.11$\pm$29.00} \\
\midrule
\multirow{3}{*}{20\%}
 & Proposed AE   & \textbf{0.071$\pm$0.051} & \textbf{0.155$\pm$0.086} & 78.38$\pm$97.49 & 96.31$\pm$128.78 & \textbf{536.0$\pm$607.5} & \textbf{618.1$\pm$708.1} & 11.13$\pm$30.24 & 15.68$\pm$40.51 \\
 & TCN-AE        & 0.182$\pm$0.138 & 0.273$\pm$0.173 & 448.65$\pm$351.88 & 475.99$\pm$373.90 & 2614.4$\pm$3103.2 & 2786.7$\pm$3163.7 & 26.02$\pm$39.34 & 29.78$\pm$42.68 \\
 & Cubic Spline  & 2.292$\pm$47.265 & 8.695$\pm$192.456 & \textbf{71.94$\pm$413.34} & \textbf{89.00$\pm$508.40} & 8188.7$\pm$228260.7 & 12167.3$\pm$342046.4 & \textbf{5.39$\pm$26.03} & \textbf{6.61$\pm$30.42} \\
\midrule
\multirow{3}{*}{30\%}
 & Proposed AE   & \textbf{0.085$\pm$0.052} & \textbf{0.173$\pm$0.083} & \textbf{113.39$\pm$115.90} & \textbf{136.19$\pm$144.42} & \textbf{837.9$\pm$778.8} & \textbf{970.9$\pm$908.9} & 11.05$\pm$25.90 & 16.61$\pm$37.84 \\
 & TCN-AE        & 0.182$\pm$0.130 & 0.274$\pm$0.163 & 445.27$\pm$314.57 & 481.44$\pm$342.87 & 2665.7$\pm$3014.2 & 2918.8$\pm$3113.2 & 25.74$\pm$36.76 & 30.85$\pm$41.68 \\
 & Cubic Spline  & 98.84$\pm$2127.51 & 413.28$\pm$8912.45 & 113.71$\pm$544.33 & 138.54$\pm$653.75 & 7096.1$\pm$148514.0 & 10474.7$\pm$223286.5 & \textbf{8.34$\pm$41.67} & \textbf{10.09$\pm$48.61} \\
\midrule
\multirow{3}{*}{40\%}
 & Proposed AE   & \textbf{0.098$\pm$0.054} & \textbf{0.190$\pm$0.087} & \textbf{148.68$\pm$137.19} & \textbf{176.72$\pm$166.19} & \textbf{1171.7$\pm$901.1} & \textbf{1366.5$\pm$1049.2} & \textbf{11.31$\pm$23.91} & \textbf{17.88$\pm$36.77} \\
 & TCN-AE        & 0.184$\pm$0.121 & 0.280$\pm$0.152 & 456.54$\pm$293.26 & 500.39$\pm$325.46 & 2707.1$\pm$2849.3 & 3040.2$\pm$2994.7 & 25.95$\pm$34.25 & 32.49$\pm$40.71 \\
 & Cubic Spline  & 134.92$\pm$2442.23 & 569.06$\pm$10342.57 & 152.93$\pm$568.22 & 188.51$\pm$702.77 & 17331.3$\pm$346503.6 & 25831.0$\pm$521194.5 & 18.88$\pm$115.52 & 22.47$\pm$135.04 \\
\midrule
\multirow{3}{*}{50\%}
 & Proposed AE   & \textbf{0.113$\pm$0.056} & \textbf{0.207$\pm$0.088} & \textbf{188.82$\pm$163.01} & \textbf{221.51$\pm$192.09} & \textbf{1573.3$\pm$1075.0} & \textbf{1826.9$\pm$1219.4} & \textbf{11.96$\pm$23.33} & \textbf{19.46$\pm$36.50} \\
 & TCN-AE        & 0.190$\pm$0.112 & 0.290$\pm$0.143 & 475.78$\pm$280.28 & 526.10$\pm$313.88 & 2833.5$\pm$2656.2 & 3268.1$\pm$2851.4 & 26.32$\pm$31.97 & 34.62$\pm$39.63 \\
 & Cubic Spline  & 412.19$\pm$8570.17 & 1738.88$\pm$36214.43 & 225.21$\pm$770.90 & 274.13$\pm$936.23 & 9566.0$\pm$181692.7 & 13843.3$\pm$273457.7 & 30.66$\pm$319.73 & 36.31$\pm$372.89 \\
\midrule
\multirow{3}{*}{Real (13.5\%)}
 & Proposed AE   & \textbf{0.063$\pm$0.053} & \textbf{0.143$\pm$0.091} & 59.70$\pm$100.04 & 72.53$\pm$128.67 & \textbf{337.0$\pm$448.6} & \textbf{388.0$\pm$528.6} & 11.41$\pm$34.94 & 15.00$\pm$43.30 \\
 & TCN-AE        & 0.183$\pm$0.143 & 0.274$\pm$0.181 & 453.48$\pm$383.19 & 473.98$\pm$400.08 & 2542.8$\pm$3119.3 & 2657.8$\pm$3156.1 & 26.43$\pm$41.50 & 29.21$\pm$43.79 \\
 & Cubic Spline  & 2.025$\pm$45.432 & 8.138$\pm$189.620 & \textbf{47.10$\pm$273.25} & \textbf{58.76$\pm$361.98} & 661.7$\pm$9456.8 & 884.7$\pm$13787.7 & \textbf{3.19$\pm$13.27} & \textbf{3.98$\pm$16.06} \\
\bottomrule
\end{tabular}}\label{Missing_Rate}
{\scriptsize Bold indicates the best-performing method for each row.\par}
\end{table*}

\subsection{Emissions inventory}
This section presents the results of the proposed approach for west-coast domestic flights in January 2025. Table~\ref{tab:emissions_by_altitude} reports fuel consumption and emissions by altitude band, together with each pollutant's share of the fleet-wide total. 

Cruise accounts for the largest absolute share of {$\mathrm{NO}_x$} mass (56.5\%), but this share is actually below cruise's share of fuel burned (64.8\%). In other words, cruise dominates in absolute terms simply because it accounts for the large majority of fuel burned overall, not because it is {$\mathrm{NO}_x$}-intensive per kilogram of fuel. {$\mathrm{NO}_x$} is instead disproportionately concentrated below cruise, as its share exceeds the corresponding fuel share at every band from LTO through high altitude, consistent with per-waypoint emission indices that peak during climb and are lowest at cruise, reflecting {$\mathrm{NO}_x$}'s thermally-driven formation, which is caused by the sustained high thrust of climb rather than cruise's lower, steadier power setting. The CO and HC values reported here for the LTO phase may be underestimated. This could happen because LTO-cycle filter misses some taxi and idle operations, or because the BFFM2 correction is less accurate at low thrust settings. However, CO and HC contribute to a small share of the total monetized air-quality cost, so this does not meaningfully change the study's main results.

\begin{figure}[H]
  \centering
  \includegraphics[width=1\linewidth]{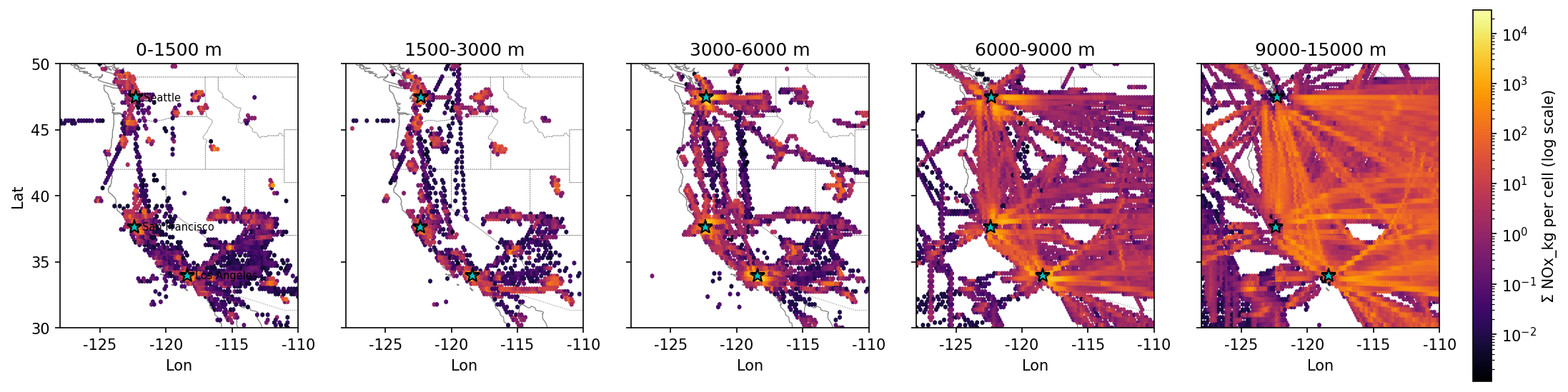}
  \caption{{$\mathrm{NO}_x$} by altitude in the US west coast area.}\label{fig:NOx_by_Altitude}
\end{figure}

\begin{figure}[htbp]
  \centering
  \includegraphics[width=1\linewidth]{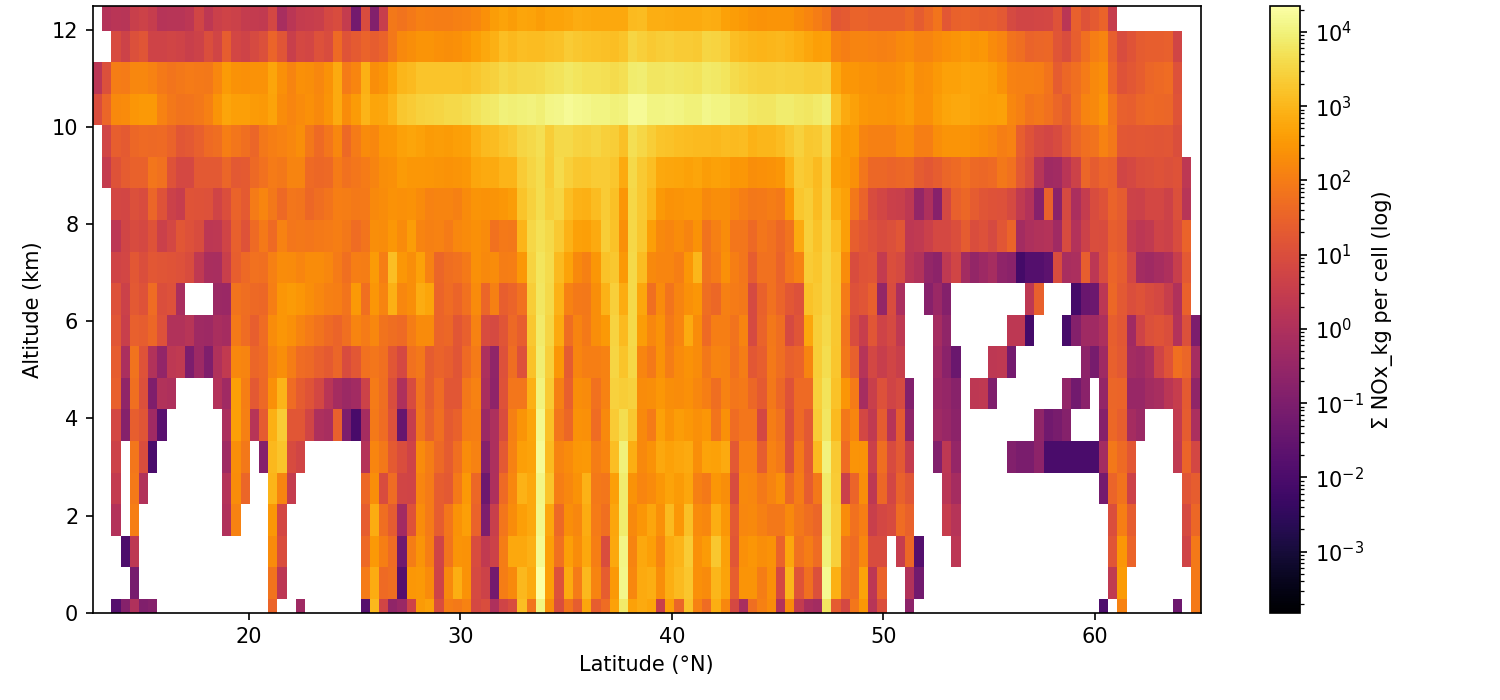}
  \caption{{$\mathrm{NO}_x$} by latitude and longitude across altitudes in the US west coast area.}\label{fig:NOx_by_lat_lon}
\end{figure}

Analyzing {$\mathrm{NO}_x$} by altitude is important since it is a species that affects both the environment and human health. Results in Figure~\ref{fig:NOx_by_Altitude} demonstrate that {$\mathrm{NO}_x$} is sparse and hub-centered at lower altitudes but spread into dense, corridor-shaped patterns at cruise altitude. This reflects {$\mathrm{NO}_x$}'s dual relevance: hub-proximate emissions at low altitude affect local air quality near airports, causing health impacts, while corridor-wide cruise-altitude emissions drive competing ozone-formation and methane-destruction pathways whose net climate effect is small and slightly net-cooling. Additionally, Figure~\ref{fig:NOx_by_lat_lon} shows the latitude-altitude distribution of {$\mathrm{NO}_x$} emissions complementary to the longitude-latitude emissions map in Figure~\ref{fig:NOx_by_Altitude}. The distribution shows a broad {$\mathrm{NO}_x$} concentration band that persist across the full latitude range at cruise altitude, while at lower altitudes emissions are more sharply concentrated near hubs at $\mathrm{33^\circ\!-\!48^\circ N}$.

\subsection{Social costs and health impact }
The social costs of emissions species based on their impacts on climate, air quality, and human health are monetized, totaling \$73.5 million (67.9\%) and \$34.7 million (32.1\%) of total social cost, respectively, in Table~\ref{tab:social_cost}, and serve as a decision-support reference for identifying where mitigation is most cost-effective. {$\mathrm{NO}_x$}'s direct climate effect is a small net cooling. $\mathrm{NO}_x$-driven destruction of atmospheric methane, a potent greenhouse gas, outweighs the short-lived warming from the tropospheric ozone $\mathrm{NO}_x$ also forms, so it acts as a modest offset to $\mathrm{CO}_2$'s climate cost rather than an additional one.

Within the air quality and health category, $\mathrm{NO}_x$ (including cruise-phase emissions, priced per \citet{grobler2019marginal}) accounts for nearly all monetized cost, at 99.7\% of the air quality and health subtotal. Accordingly, a {$\mathrm{NO}_x$}-specific pricing scheme as a complementary fee to carbon pricing would offer a sharper and more effective tool to reduce aviation's environmental and near-airport health impact compared to carbon pricing alone.

\begin{table}[htbp]
\centering
\footnotesize
\setlength{\tabcolsep}{4pt}
\renewcommand{\arraystretch}{0.85}
\caption{Fuel burn and emissions by altitude, west coast January-2025 inventory.}
\label{tab:emissions_by_altitude}
\resizebox{\linewidth}{!}{%
\begin{tabular}{lrrrrrrcccccc}
\toprule
& Fuel (t) & \multicolumn{5}{c}{Total (kg)} & \multicolumn{6}{c}{Share of fleet total} \\
\cmidrule(lr){2-2} \cmidrule(lr){3-7} \cmidrule(lr){8-13}
Altitude band & Fuel & $\mathrm{CO}_2$ & $\mathrm{NO}_x$ & $\mathrm{CO}$ & HC & nvPM & Fuel & $\mathrm{CO}_2$ & $\mathrm{NO}_x$ & $\mathrm{CO}$ & HC & nvPM \\
\midrule
LTO ($<$914 m) & 6{,}051.0 & 19{,}121{,}232.7 & 92{,}908.8 & 6{,}915.9 & 194.1 & 201.2 & 4.9\% & 4.9\% & 6.3\% & 17.5\% & 4.9\% & 3.7\% \\
Low (914--3 km) & 8{,}937.9 & 28{,}243{,}764.0 & 151{,}073.6 & 1{,}948.8 & 197.2 & 388.6 & 7.2\% & 7.2\% & 10.2\% & 4.9\% & 4.9\% & 7.1\% \\
Mid (3--6 km) & 12{,}286.6 & 38{,}825{,}561.2 & 187{,}137.9 & 2{,}549.3 & 299.9 & 540.6 & 9.9\% & 9.9\% & 12.7\% & 6.5\% & 7.5\% & 9.9\% \\
High (6--9 km) & 16{,}656.6 & 52{,}634{,}856.0 & 210{,}635.8 & 4{,}104.1 & 482.8 & 732.9 & 13.4\% & 13.4\% & 14.3\% & 10.4\% & 12.1\% & 13.4\% \\
Cruise ($>$9 km) & 80{,}755.8 & 255{,}188{,}454.4 & 834{,}213.4 & 24{,}025.1 & 2{,}826.5 & 3{,}591.1 & 64.8\% & 64.8\% & 56.5\% & 60.8\% & 70.7\% & 65.8\% \\
\midrule
Total & 124{,}687.9 & 394{,}013{,}868.3 & 1{,}475{,}969.5 & 39{,}543.1 & 4{,}000.6 & 5{,}454.4 & 100\% & 100\% & 100\% & 100\% & 100\% & 100\% \\
\bottomrule
\end{tabular}}
\vspace{2pt}
\end{table}

\begin{table}[htbp]
\centering
\caption{Social cost of west coast January-2025 aviation emissions.}
\label{tab:social_cost}
\renewcommand{\arraystretch}{0.85}
\resizebox{\linewidth}{!}{%
\footnotesize
\begin{tabular}{llrrr}
\toprule
Category & Pollutant & Tons & Unit price (USD/t) & Cost (USD) \\
\midrule
\multirow{5}{*}{Climate (all altitudes, whole area)}
 & $\mathrm{CO}_2$   & 394{,}013.9 & 190      & 74{,}862{,}635 \\
 & $\mathrm{NO}_x$   & 1{,}476.0   & $-$918 & $-$1{,}354{,}893 \\
 & HC                & 4.0         & 0        & 0 \\
 & $\mathrm{CO}$     & 39.5        & 0        & 0 \\
 & $\mathrm{nvPM}$   & 5.5 & 0 & 0 \\
\cmidrule(lr){2-5}
 & \textit{Subtotal} & & & \textbf{73{,}507{,}742} \\
\midrule
\multirow{5}{*}{\shortstack[l]{Air quality \& Human health\\(LTO $\leq$914 m, $\leq$50 km of KLAX/KSFO/KSEA,\\ plus cruise-phase $\mathrm{NO}_x$; USA-specific pricing)}}
 & $\mathrm{NO}_x$ (LTO)     & 72.3   & 20{,}000  & 1{,}446{,}000 \\
 & $\mathrm{NO}_x$ (cruise) & 1{,}383.1 & 24{,}000  & 33{,}193{,}457 \\
 & $\mathrm{PM}_{2.5}$ (primary $\mathrm{nvPM}$) & 0.176  & 600{,}000 & 105{,}600 \\
 & $\mathrm{CO}$     & 2.154  & 520       & 1{,}120 \\
 & HC                & 0.105  & 19{,}000  & 1{,}995 \\
\cmidrule(lr){2-5}
 & \textit{Subtotal} & & & \textbf{34{,}748{,}172} \\
\midrule
\multicolumn{4}{l}{\textbf{Total social cost}} & \textbf{108{,}255{,}914} \\
\bottomrule
\end{tabular}}
\vspace{2pt}

\end{table}

\section{Discussion and conclusion}
Paving the path toward sustainable air transportation sector is crucial due to increasing demand and lagging behind other transportation sectors, because of its damage through climate, air quality and human health. Since aviation technologies emerge by a slower pace compare to other sectors, for instance electric and Automated Connected Vehicles (CAVs), thus the near-term solutions would be even more practical. One focused solution is policies; Yet, to have realistic, and efficient policies we need to have accurate data for both flights and emissions. This work presents an approach for providing a 3-D emissions map using ADS-B flight trajectories imputed by the proposed hybrid AE model for the purpose of mapping social and health costs for the domestic flights in the west coast area in the United States. 

The results show that $\mathrm{CO_2}$ emissions contribute \$74.9~million to climate damage, while $\mathrm{NO_x}$ emissions provide a small net climate credit of \$1.35~million but contribute \$34.6~million to air-quality and health impacts, where the majority is from cruise phase emissions, making $\mathrm{NO_x}$ by far the largest driver of the inventory's monetized health cost. $\mathrm{HC~(NMVOC)}$ and CO health costs are comparatively small, \$1{,}995 and \$1{,}120, respectively, and together with $\mathrm{nvPM}$ (valued as primary $\mathrm{PM_{2.5}}$, \$105{,}600) correspond to an estimated 0.002 premature deaths across the three hub areas for the study length, derived from primary $\mathrm{PM}_{2.5}$ mass alone (0.0128 deaths/tonne, \$7.4M VSL, 2006\$\cite{epa_mortality_risk_valuation}), independent of the \citet{grobler2019marginal}-based costs for $\mathrm{NO}_x$/HC/CO and excluding $\mathrm{NO}_x$-driven secondary PM$_{2.5}$ mortality. Thus, this is not an aggregate mortality estimate. The results are only credible as the model could use complete accurate ADS-B trajectories, which is achieved in this paper using the proposed CBiG3D-AE model ((CNN-Bi-GRU + 3D-CNN) + (TCN) AE) that outperforms both a vanilla TCN-AE and a Cubic Spline interpolation method. Imputation accuracy directly affects the quality of the emissions inventory and climate and social costs estimations. 

The proposed approach outputs a high-resolution emissions inventory associated with climate and social cost, which is the key for understanding the aviation impacts on the climate, air quality and human health and evaluating their costs reduction alternatives. This research infers the suitability of using imputation models for ADS-B data to use for an accurate emissions inventory implementation to provide a climate and social cost estimation that could be taken into account by decision-makers and policy-makers through their paths forward to emissions and their social costs reductions. For future work, we will consider emissions species costs per weight based on their altitudes as a dynamic cost function and also consider a larger area in the US to study domestic flights emissions and costs.

\section{Acknowledgments}

This document has been created with or contains elements of Base of Aircraft Data (BADA) Family 3 Release 3.16 which has been made available by EUROCONTROL to the University of Washington. EUROCONTROL has all relevant rights to BADA. © 2026 EUROCONTROL - European Organization for the Safety of Air Navigation \cite{eurocontrol}. All rights reserved. EUROCONTROL shall not be liable for any direct, indirect, incidental or consequential damages arising out of or in connection with this product or document, including with respect to the use of BADA. We appreciate OpenSky Network website for their assistance in providing ADS-B flight trajectories. The Claude AI is used to implement and edit code, and to review the manuscript for grammatical errors.

\section*{AUTHOR CONTRIBUTIONS}
The authors confirm contribution to the paper as follows: 
study conception and design: Hesam Shafiei Nia, Don MacKenzie; 
data collection: Hesam Shafiei Nia; 
analysis and interpretation of results: Hesam Shafiei Nia; 
draft manuscript preparation: Hesam Shafiei Nia, Don MacKenzie. 
All authors reviewed the results and approved the final version of the manuscript.

\section*{DECLARATION OF CONFLICTING INTERESTS}
All authors declare no potential conflicts of interest with respect to the research, authorship, and publication of this article.

\section*{FUNDING}
The authors disclosed no financial support for the research, authorship, and/or publication of this article.

\bibliographystyle{trb}
\bibliography{trb_template}
\end{document}